\documentclass[10pt,twocolumn,letterpaper]{article}

\usepackage{cvpr}              

\usepackage[dvipsnames]{xcolor}

\usepackage{multirow}
\usepackage{xcolor}
\usepackage{subcaption}
\usepackage{wrapfig}
\usepackage{comment}
\usepackage{tikz,standalone,pgfplots}
\usepackage{url}
\pgfplotsset{compat=1.18} 

\def\treadmill{Treadmill Dataset}
\def\outdoor{Outdoor Dataset}
\def\earlyfusion{early fusion}
\def\modelfusion{model fusion}

\def\Imageonly{Image-only}
\def\EarlyFusion{Early-fusion}
\def\ModelFusion{Model-fusion}

\definecolor{cvprblue}{rgb}{0.21,0.49,0.74}
\usepackage[pagebackref,breaklinks,colorlinks,citecolor=cvprblue]{hyperref}

\def\paperID{10} 
\def\confName{CV4Animals}
\def\confYear{2024}

\title{CLHOP: Combined Audio-Video Learning for Horse \\ 3D Pose and Shape Estimation}
\author{Ci Li$^1$ ~~~~ Elin Hernlund$^2$ ~~~~ Hedvig Kjellström$^{1,2}$ ~~~~ Silvia Zuffi$^3$\\
$^1$ KTH, Sweden ~~~~ $^2$ SLU, Sweden ~~~~ $^3$ IMATI-CNR, Italy}

\begin{document}
\twocolumn[{%
\renewcommand\twocolumn[1][]{#1}%
\maketitle
\begin{center}
\vspace{-8mm}
    \centering
    \includegraphics[width=0.8\textwidth]{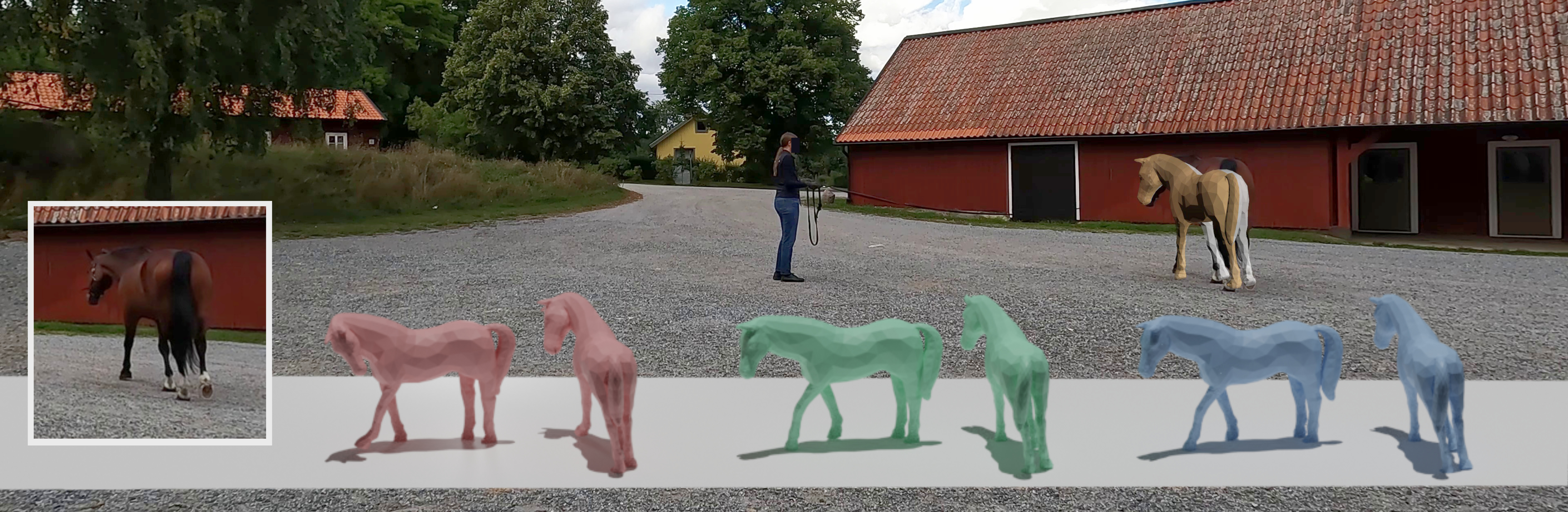}
    \captionof{figure}{We estimate the articulated 3D motion of a horse from video, combining both visual and auditory information. We show that by training with both of these modalities, we are able to reconstruct poses that are more accurate and natural, even under self-occlusion. 
    In figure, we show results for an Image-only network ({\color{red}red}) and for two networks that exploit audio: Early-fusion ({\color{green}green}) and Model-fusion ({\color{blue}blue}). Note that both audio-based networks can reconstruct more natural head pose and correctly estimate the front left hoof touching the ground.}
    \label{fig:front}
\end{center}
\vspace{-3mm}
}]
\begin{abstract}
\vspace{-3mm}
In the monocular setting, predicting 3D pose and shape of animals typically relies solely on visual information, which is highly under-constrained. In this work, we explore using audio to enhance 3D shape and motion recovery of horses from monocular video. We test our approach on two datasets: an indoor treadmill dataset for 3D evaluation and an outdoor dataset capturing diverse horse movements, the latter being a contribution to this study. Our results show that incorporating sound with visual data leads to more accurate and robust motion regression. This study is the first to investigate audio's role in 3D animal motion recovery.
\end{abstract}    
\vspace{-3mm}
\section{Introduction}
\vspace{-2mm}
Advancements in computer vision and machine learning have greatly propelled 3D markerless motion capture of humans and animals. Notably, using parametric models, like the SMPL model~\cite{loper2015smpl} for humans and the SMAL model~\cite{zuffi20173d} for quadrupeds, has pushed this research area forward. These methods infer subject motions solely from monocular images or videos \cite{bogo2016keep,kanazawa2018end,kolotouros2019learning,VIBE:CVPR:2020,Ruegg:AAAI:2020,zhang2020object,Kocabas_SPEC_2021,humanMotionKanazawa19,biggs2018creatures, zuffi2019three, biggs2020left, ruegg2022barc, ruegg2023bite}. However, the integration of multimodal data in this context remains underexplored.

Human perception combines senses like vision, crucial for understanding object movement, and hearing, which complements vision and enhances our comprehension of the environment. Prior research highlights the synergy between sound and visual data in motion estimation and animation \cite{shlizerman2018audio,li2021ai,li2022danceformer,kucherenko2019analyzing}. This research exploits the correlation of audio and visual data for capturing articulated 3D motion from monocular videos, specifically for horses.

Horses play a significant role in various human activities. The need for advanced markerless motion capture techniques to analyze equine behavior and health is growing~\cite{ani13030390}. Their unique sounds through their hooves and respiratory actions, provide rich audio cues for motion analysis. We are the first to combine visual and audio data for 3D animal motion reconstruction. 

Our study necessitates a dataset with both audio and video data. With only \textit{the \treadmill}~\cite{rhodin2018vertical} available, we introduce \textit{the \outdoor}, comprising four horses on an outdoor gravel surface, recorded with a 4K camera and synchronized audio. This dataset broadens this field by offering diverse motion and audio for 3D motion modeling.

We develop two fusion strategies
for accurate shape and pose estimation via the hSMAL model~\cite{li2021hsmal}, a horse-variant of SMAL:
1) \textit{\earlyfusion} that employs audio data both in training and testing and 2) \textit{\modelfusion} that only leverages audio information during the training phase.
Experiments with different setups show how audio information facilitates 3D horse reconstruction learning.
We demonstrate that the networks learning with audio data using our fusion strategies are more robust to changes in appearance and visual ambiguities, as illustrated in (Fig. \ref{fig:front}).
\vspace{-2mm}
\section{Related Work}
\vspace{-1mm}
\paragraph{Model-Based 3D Pose Estimation From Images} Monocular markerless motion capture of articulated subjects like humans and animals relies on prior models of body shape and pose. This literature review focuses on the use of SMPL model~\cite{loper2015smpl}, relevant to our horse model approach, pivotal for estimating human pose from visuals~\cite{bogo2016keep, kanazawa2018end, kolotouros2019learning, VIBE:CVPR:2020, Ruegg:AAAI:2020, zhang2020object,STRAPS2020BMVC, Kocabas_SPEC_2021, humanMotionKanazawa19}, addressing interactions with environment or objects~\cite{rempe2021humor, shimada2020physcap, zhang2020phosa}, multiple bodies in scenes~\cite{Kocabas_PARE_2021, ROMP:ICCV:2021, jiang2020coherent}, and camera distortions~\cite{Kocabas_SPEC_2021}. 
Model-based methods have been applied to specific animal species, including birds~\cite{badger20203d, wang2021birds} quadrupeds~\cite{zuffi20173d}, zebras~\cite{zuffi2019three}, dogs~\cite{Kearney_2020_CVPR,biggs2018creatures,biggs2020left, ruegg2022barc, ruegg2023bite} and horses~\cite{li2021hsmal}. 
None of these methods incorporate multimodal data. 

\vspace{-4mm}
\paragraph{Pose Estimation and Synthesis From Audio} Audio-driven research shows a strong link between sound and motion. 
Studies have mapped speech to facial movement in 3D~\cite{cudeiro2019capture, richard2021meshtalk} 
or animated faces with realistic expressions from speech in 2D portraits~\cite{zhou2020makelttalk}. 
Other works predict upper body movements from instrument music \cite{shlizerman2018audio}, convert speech into gestures \cite{kucherenko2019analyzing}, synchronize 3D body gestures and facial expressions with speech \cite{3dconvgesture_2021}, generate dance movements from music \cite{valle2021transflower,li2021ai}. All these methods demonstrate audio's role for complex animations. However, animal motion analysis with audio, like horse behavior detection \cite{nunes2021horse} and primate action recognition \cite{doi:10.1126/sciadv.abi4883}, is less explored, presenting a research opportunity.

\vspace{-4mm}
\paragraph{Multimodality Fusion} Recent studies explore combining audio and visual data for integrated representations in multiple applications, like speech separation~\cite{owens2018audio}, egocentric action recognition~\cite{kazakos2019epic}, scene understanding~\cite{gao2021visualvoice}, which need both modalities present in the inference stage. 
Further research addresses missing modalities at inference, with a cycle translation training \cite{pham2019found},  
an alignment of independent latent spaces for multimodality data~\cite{yadav2021speech}, an optimization of joint representation that separates shared and specific modal factors~\cite{tsai2019learning}, a utilization of Bayesian networks and meta-learning~\cite{ma2021smil}. Variational AutoEncoders (VAEs) are key techniques with different variants~\cite{suzuki2016joint,yadav2020bridged,wu2018multimodal,shi2019variational} in multimodal learning. 

The use of multimodal data in 3D pose estimation remains unexplored. Yang et al.'s study stands out as a rare example, focusing on human 3D pose estimation with metric scale~\cite{yang2022posekernellifter}. Instead, we focus on modeling the 3D mesh of the animals. Drawing inspiration from research on emotion detection through multimodal data integration~\cite{han2019implicit, han2019emobed}, we follow a similar path that combines video and audio for estimating the 3D motions of animals. 
\vspace{-2mm}
\section{Method}
\vspace{-1mm}
\begin{figure}[tb]
\centering
    \vspace{-4mm}
    \includegraphics[width=\linewidth]{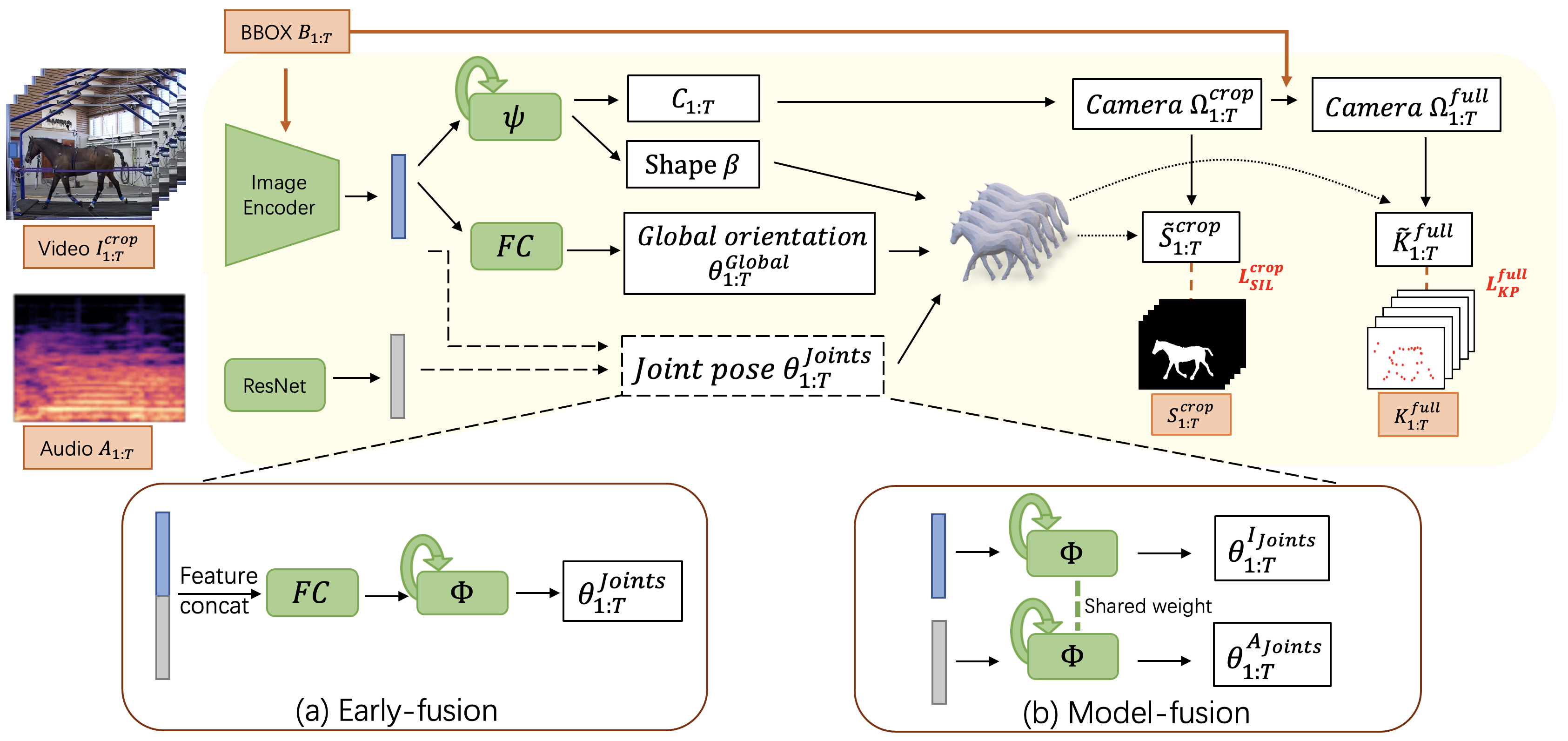}
    \vspace{-6mm}
    \caption{Video--audio fusion frameworks. (a) \EarlyFusion{}, (b) \ModelFusion{}. Both networks use the same architecture for feature extraction and predicting {\small $C_{1:T}$, $\beta$, $\theta_{1:T}^{Global}$} using video features. The key difference lies in estimating $\theta_{1:T}^{Joints}$,: (a) combines video and audio features before $\Phi$ to estimate $\theta_{1:T}^{Joints}$ and (b) processes through the shared $\Phi$ separately to obtain $\theta_{1:T}^{I_{Joints}}$ and $\theta_{1:T}^{A_{Joints}}$. All parameters are then mapped to 3D meshes $\mathbf{v_{1:T}}$ for 2D projection and loss calculation. Inputs are in \textcolor{orange}{orange} and learnable modules in \textcolor{green}{green}.
    }
    \label{fig:two_frameworks}
    \vspace{-5mm}
\end{figure}
In this section, we first introduce the hSMAL model~\cite{li2021hsmal} utilized in this work. Then, we propose a backbone module for video data processing, followed by two fusion strategies to integrate auxiliary audio data into the learning and inference process. Finally, we present how we construct the loss function for learning.

\vspace{-4mm}
\paragraph{The hSMAL Model} We reconstruct 3D horses from monocular video and audio inputs by estimating the hSMAL model parameters~\cite{li2021hsmal}, a horse-adapted version of SMAL~\cite{zuffi20173d} that defines the shape and pose of horses. The model uses a mapping {\footnotesize $\Theta : (\beta, \theta) \mapsto \mathbf{v}$}, where $\beta$ are PCA coefficients for the model's shape space,  {\footnotesize $\theta=(\theta^{Global}, \theta^{Joints})$} indicates the global orientation and joint rotations and $\mathbf{v}$ represents the 3D mesh vertices. 

\vspace{-4mm}
\paragraph{Fusion Strategies} Our two fusion strategies (\textbf{\earlyfusion{}} and \textbf{\modelfusion{}})
are both based on our proposed backbone module denoted as \textbf{\Imageonly{} network}.  \Imageonly{} network adapts the CLIFF architecture~\cite{li2022cliff} for video sequences, estimating 3D pose and shape by considering the subject's location in the full image frame. This approach is particularly effective for our outdoor data, where horses move at varying distances from the camera.

\vspace{-4mm}
\paragraph{\Imageonly{} network} 
consists of an image encoder processing a video clip $I^{crop}_{1:T}$ and bounding box data {\footnotesize $B_{1:T}=\left \{ B_t \right \}_{t=1}^{T}$} of $T$ frames, with a ResNet backbone and a temporal encoder as final visual features. Each $B_t$ at time $t$ is defined as {\footnotesize $B_t=\left [ \frac{c_{x_t}}{f_{full}}, \frac{c_{y_t}}{f_{full}}, \frac{b_t}{f_{full}} \right ]$} with the center ($c_{x_t}$, $c_{y_t}$) and size $b_t$ of the original bounding box and the focal length $f_{full}$ for the original camera $\Omega^{full}$, calculated as {\footnotesize $f_{full}=\sqrt{w^2+ h^2}$} with the original image's width $w$ and height $h$, assuming a $50^{\circ}$ diagonal Field-of-View~\cite{kissos2020beyond}.

To predict parameters, we employ an iterative error feedback (IEF) loop, similar to \cite{kanazawa2018end,li2022cliff}, called the \textit{Regression Block}. Here, $\psi$ predicts shape and camera parameters, while $\Phi$ focuses on pose estimation.

Block $\psi$ processes visual features to estimate model shape $\beta$ and camera {\footnotesize $\Omega^{crop}_{1:T}=(f_{crop}, \Gamma^{crop}_{1:T})$}, with weak perspective projection parameters $C_{1:T}$ including scale $s_{1:T}$ and translation ($p_{x_{1:T}}$, $p_{y_{1:T}}$). The full camera translation {\footnotesize $\Gamma^{crop}_{1:T}= \left \{ \Gamma^{crop}_t \right \}_{t=1}^{T}$} for the cropped image is calculated as {\footnotesize $\Gamma^{crop}_{t} = \left[ p_{x_{t}}, p_{y_{t}}, \frac{2 \cdot f_{crop}}{r \cdot s_{t}} \right]$} with {\footnotesize $f_{crop} = 5000$} is the predefined focal length of the camera $\Omega^{crop}$ and $r = 224$ is the bounding box size. 
We convert the cropped camera {\small $\Omega^{crop}_{1:T}$} to the original camera {\footnotesize $\Omega^{full}_{1:T}=(f_{full},\Gamma^{full}_{1:T})$},  for reprojecting 3D points to the full image, with the translation calculated as {\footnotesize $\Gamma^{full}_{t} = \left[ p_{x_{t}} + \frac{2 \cdot c_{{x}_{t}}}{b_{t} \cdot s_{t}}, p_{y_{t}} + \frac{2 \cdot c_{{y}_{t}}}{b_{t} \cdot s_{t}}, \frac{2 \cdot f_{full}}{b_{t} \cdot s_{t}} \right]$}.

For pose $\theta$ estimation, the global rotation {\small $\theta^{Global}_{1:T}$} is estimated directly from visual features via a fully-connected layer, eliminating manual global rotation initialization. The pose parameters $\theta^{Joints}_{1:T}$ are estimated by Block $\Phi$ using visual features as input.

Regarding the fusion strategies, the main differences are the given information as the inputs to $\Phi$ for 3D pose estimation. Audio features are derived by converting audio $A_{1:T}$ into a log-mel spectrogram with Librosa~\cite{mcfee2015librosa}, then using a ResNet backbone for feature extraction.

In \textbf{\earlyfusion{} strategy} (Fig.\ref{fig:two_frameworks}.a), the encoded visual and audio features are concatenated and processed through two FC layers before entering $\Phi$ to predict the pose parameters $\theta^{Joints}_{1:T}$, forming the \EarlyFusion{} network. This method requires video and audio data during inference.

In \textbf{\modelfusion{} strategy} (Fig.\ref{fig:two_frameworks}.b), visual and audio features are fed separately into $\Phi$, producing two sets of pose parameters: {\small $\theta^{I_{Joints}}_{1:T}$} from visuals and {\small $\theta^{A{Joints}}_{1:T}$} from audio, defining the \ModelFusion{} network. Visual data is the primary modality, with audio as an auxiliary to enhance pose estimation accuracy and we only evaluate the poses from visual data. During inference, the model can operate with just the primary visual input, allowing for the absence of audio. 

The predicted parameters {\footnotesize $({\beta, \theta^{Global}_{1:T}, \theta^{Joints}_{1:T}})$} generate the hSMAL model {\small $\mathbf{v_{1:T}}$}. Then, the 3D vertices are projected to the original image frame to derive 2D keypoints {\small $\tilde{K}^{full}_{1:T}$}, using the original camera {\small $\Omega^{full}_{1:T}$}. Differing from CLIFF, we utilize Pytorch3D~\cite{ravi2020pytorch3d, liu_soft_2019} to render silhouettes $\tilde{S}^{crop}_{1:T}$ in the bounding box frame, employing the 3D model mesh and the cropped image camera $\Omega^{crop}_{1:T}$, avoiding original frame rendering for computational efficiency.

\vspace{-4mm}
\paragraph{Training Losses}
All regression networks are trained end-to-end with the loss defined as {\footnotesize $L = L_{KP}+L_{SIL}+L_{SMOOTH}+L_{HSMAL}$}. $L_{KP}$ and $L_{SIL}$ represent the 2D photometric loss for keypoints and silhouettes, penalizing the difference between the predicted and groundtruth values. $L_{SMOOTH}$ enhance the temporal smoothness of the predicted parameters across frames, while $L_{HSMAL}$ applies the shape and pose prior of the hSMAL model~\cite{li2021hsmal}. Check \textit{Supplementary Material} for more details.

\vspace{-2mm}
\section{Experiments}
\vspace{-1mm}
\paragraph{Datasets} The \textit{\treadmill{}}, acquired from the University of Z{\"u}rich~\cite{rhodin2018vertical}, includes video, audio, and 3D motion capture recordings of seven horses trotting on a treadmill. 2D groundtruth keypoints are created with the mocap data, supplement with DeepLabCut~\cite{Mathisetal2018,NathMathisetal2019}, and groundtruth silhouettes are from OSVOS~\cite{caelles2017one}. 
We further introduce the \textit{\outdoor{}} for evaluating our network in natural settings. Captured with a GoPro10 at 4K and synchronized audio, it records four horses (white, black, brown, red) performing walks, trots, and canters in both directions under human guidance. Groundtruth keypoints and silhouettes are obtained with ViTPose+~\cite{xu_vitpose_2022} and Detectron2~\cite{wu2019detectron2} respectively. More details are in \textit{Supplementary Material}.

\vspace{-5mm}
\paragraph{Experiment Setup} We conduct two experiments on both datasets. The first evaluates the impact of appearance variations by splitting test subjects into those with colors similar to training data (Test Data 1) and those with significantly different colors (Test Data 2), challenging the network with out-of-distribution data. 
This evaluates the audio inputs that can complement visual information when the latter is less reliable or informative. The second experiment assesses the robustness of our audio-enhanced models to visual interruptions, using synthetically occluded video frames. Examples of the synthetic occlusions are in \textit{Supplementary Material}.

\vspace{-5mm}
\paragraph{Experiments on \treadmill{}} We demonstrate results on the \treadmill{}, where we have quantitative 3D evaluation. 
\begin{figure*}[htp]
\centering
    \vspace{-4mm}
    \includegraphics[width=\linewidth]{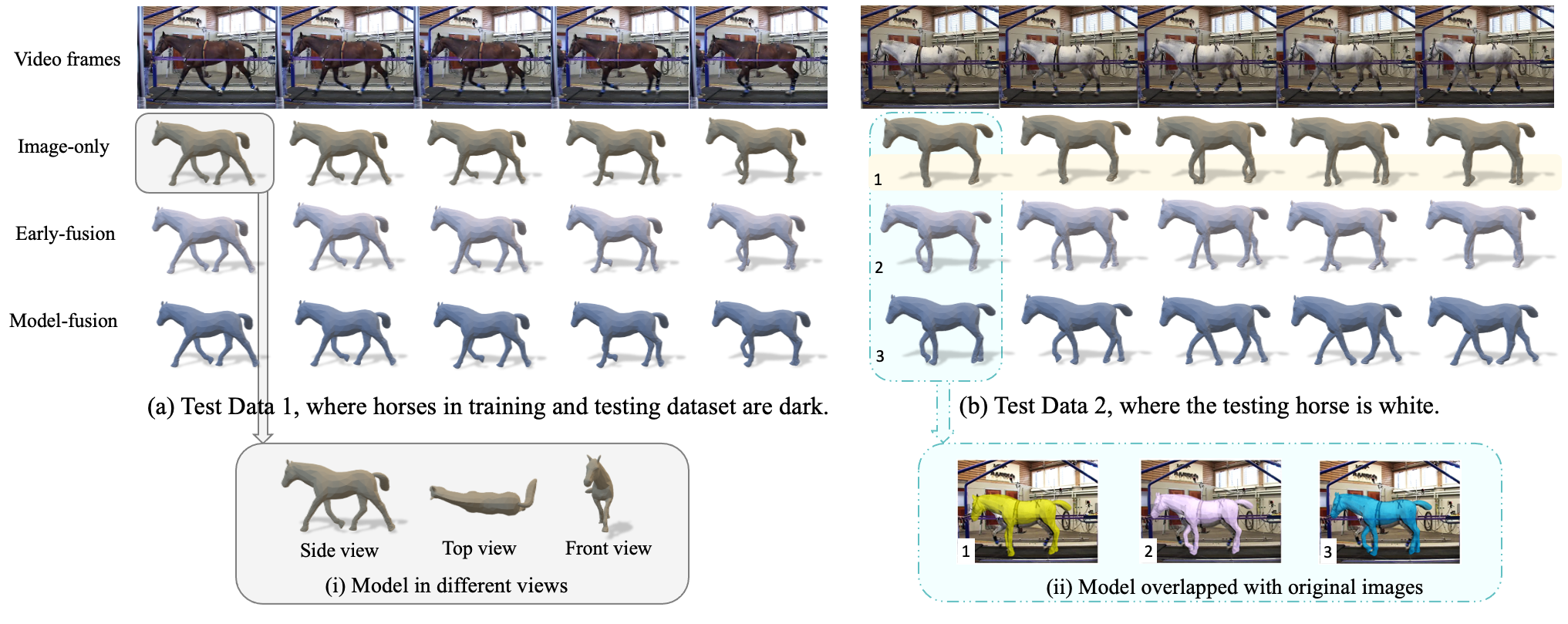}
    \vspace{-8mm}
    \caption{Example results of different networks in the Treadmill Dataset with Test Data 1 (a) and Test Data 2 (b). (i) The model is shown in different views. (ii) Model overlapped with the original images. Refer to the main text for more details.}
    \vspace{-6mm}
    \label{fig:Tread_results}
\end{figure*}
The model is trained on three randomly selected dark-colored horses, using 75\% of the recording for training and 25\% for validation. Test Data 1 comprises the remaining three dark-colored horses and Test Data 2 contains the white horse. The results are reported as the mean per 3D joint position error after rigid alignment with Procrustes analysis (P-MPJPE)~\cite{gower1975generalized}, in mm, given the accurate 3D mocap data. We report the mean and standard deviation for all networks on the Treadmill Datasets in Tab.~\ref{tab:Tread_results}. The results using the optimization method~\cite{li2021hsmal} are included, serving as an upper bound as the model uses the additional ground-truth mocap data. 

In Test Data 1, performances across all networks are comparable. The Image-only network shows similar performance to the \EarlyFusion{} network and the \ModelFusion{} network. This suggests that for Test Data 1 there is enough information in the visual modality to correctly estimate the horse motion, as the appearance of the training and test horses is similar. We perform a non-parametric Wilcoxon significant test to compare the P-MPJPE errors of the \Imageonly{} network against both the \EarlyFusion{} network and the \ModelFusion{} networks. We obtain p-values of 5.2e-19 and 1.2e-46, respectively. This shows that the differences in errors between the methods are statistically significant.
Test Data 2, the network faces a more challenging task, as the color of the test horse has not been seen during training. The performance of the \Imageonly{} network drops in Test Data 2, highlighting the difficulty posed by the large appearance difference between training and test data. Both the Early- and \ModelFusion{} networks perform better than the \Imageonly{}, with \ModelFusion{} performing the best, which shows that audio is an effective source to enhance the robustness of appearance variation even though the visual information varies.

The question remains whether a data augmentation strategy can provide comparable robustness to appearance changes. We introduce color jittering augmentation during training, adding variations in contrast, brightness, saturation, and hue. 
The performance of the Early-fusion network and Model-fusion network are slightly better than the performance achieved with data augmentation in Test Data 1 and the Model-fusion network performs better in Test Data 2. Data
augmentation reduces the train/test domain gap, but training with multimodal data gives better results, indicating that audio is an effective way to improve robustness to appearance variation.

\begin{table}[b]
\centering
\vspace{-5mm}
\setlength\tabcolsep{4pt}
\caption{Network comparison on the Treadmill Dataset (mean $\pm$ std). Unit: mm. * denotes networks with data augmentation. $\nmid$ denotes optimization with mocap and image as input.}\label{tab:Tread_results}
\vspace{-4mm}
\scalebox{0.8}{
\begin{tabular}{c|cc|cc}
\toprule
\multirow{2}{*}{P-MPJPE$\downarrow$} & \multicolumn{2}{c|}{Original Data } & \multicolumn{2}{c}{Synthetic Occluder } \\ \cline{2-5}
                        & Test Data 1         & Test Data 2         & Test Data 1         & Test Data 2         \\ \midrule
Li et al.\cite{li2021hsmal} $\nmid$ & \textit{65$\pm$7}    & \textit{69$\pm$5}   & -                   & -                   \\
Image-only$^*$                      & 83$\pm$12            & 115$\pm$53          & 136$\pm$36          & 149$\pm$40          \\ \midrule

Image-only                          & 83$\pm$13            & 135$\pm$80          & 127$\pm$41          & 146$\pm$36          \\
Early-fusion                        & 82$\pm$14            & 116$\pm$62          & 159$\pm$76          & 176$\pm$81          \\
Model-fusion                       & \textbf{82$\pm$13}   & \textbf{112$\pm$55} & \textbf{126$\pm$37} & \textbf{138$\pm$43} \\ 
\bottomrule
\end{tabular}
}
\end{table}

In the synthetic occluder experiments, the simulated occluder covers a large area of body parts of the horse. Tab.~\ref{tab:Tread_results} shows that such extreme occlusions impact all networks' performance. However, the \ModelFusion{} network outperforms all others, which confirms its robustness, while the Early-fusion network is more sensitive to noisy visual cues. 

Fig.~\ref{fig:Tread_results} shows visual examples from different networks. 
Fig.~\ref{fig:Tread_results}(i) shows that the model's tail bends to the right given that the network is only supervised by 2D information, and the horses have a braided tail, which is not represented by the hSMAL model.
In Test Data 1 (Fig.~\ref{fig:Tread_results}a), all networks that use visual features for pose estimation, like \Imageonly{}, \EarlyFusion{}, and \ModelFusion{} network, perform similarly. 
In Test Data 2 (Fig.~\ref{fig:Tread_results}b), the \Imageonly{} network often predicts rigid legs. 
\begin{wrapfigure}{r}{0.29\textwidth} 
  \centering
  \vspace{-4mm}
  \includegraphics[width=0.28\textwidth]{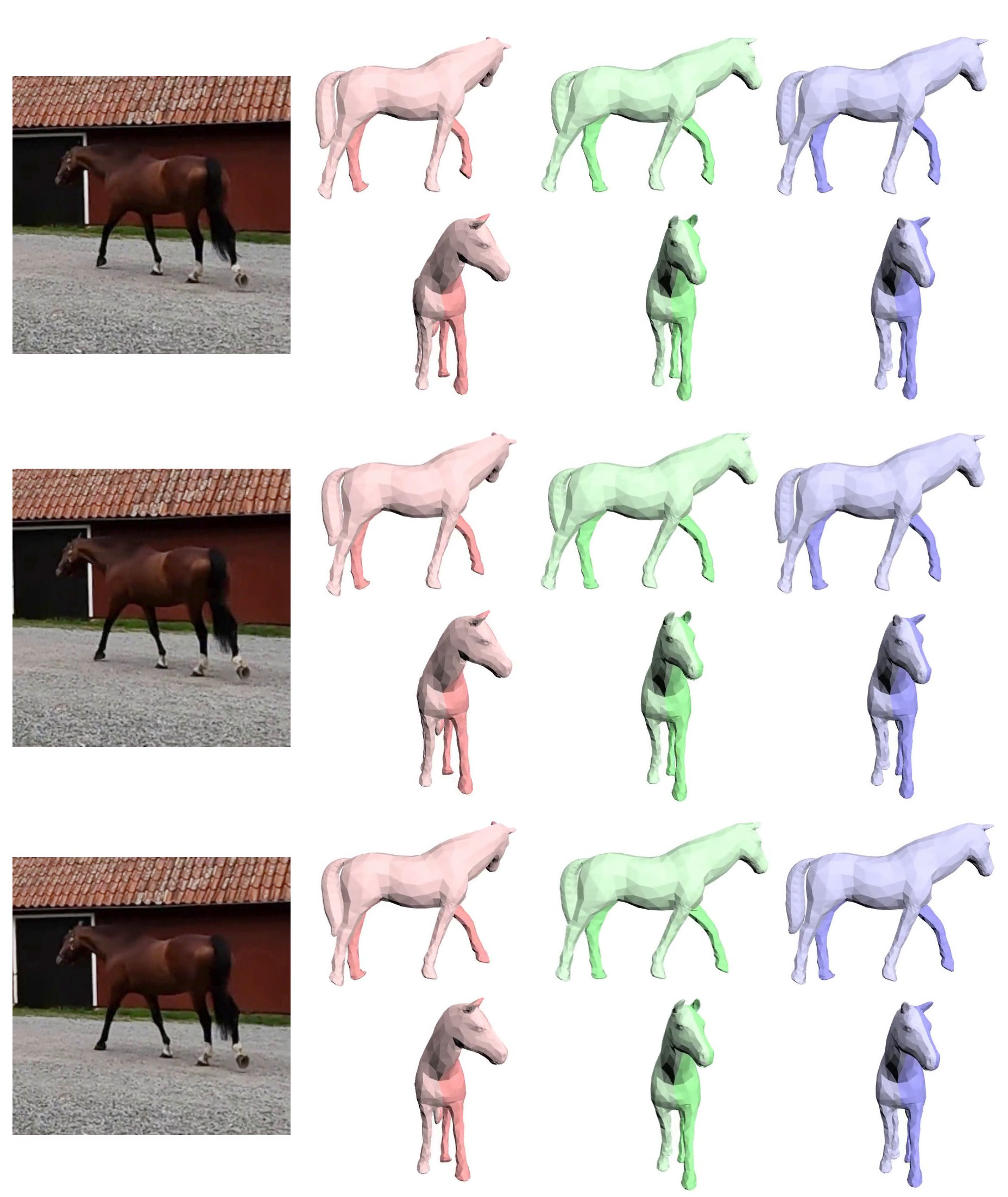} 
  \vspace{-4mm}
  \caption{Samples of full body visible for the Outdoor Dataset. Image-only Network (in \textcolor{pink}{LightRed}), \EarlyFusion{} Network (in \textcolor{green}{in LightGreen}), \ModelFusion{} Network (in \textcolor{blue}{LightBlue}).}
  \label{fig:example_H3S1_1_full_visible}
  \vspace{-5mm} 
\end{wrapfigure}
The \EarlyFusion{} network manages to estimate plausible poses for the first three frames, while the \ModelFusion{} network consistently predicts correct poses. This indicates the robustness of networks that incorporate audio in accurately estimating horse motions, especially in situations where visual features alone are insufficient or when self-occlusion occurs, as the right legs are often not fully visible.
    \vspace{-1mm} 
\begin{figure*}[ht]
    \centering
{\begin{subfigure}[b]{0.32\textwidth}
	\centering
     \includegraphics[width=1\linewidth]{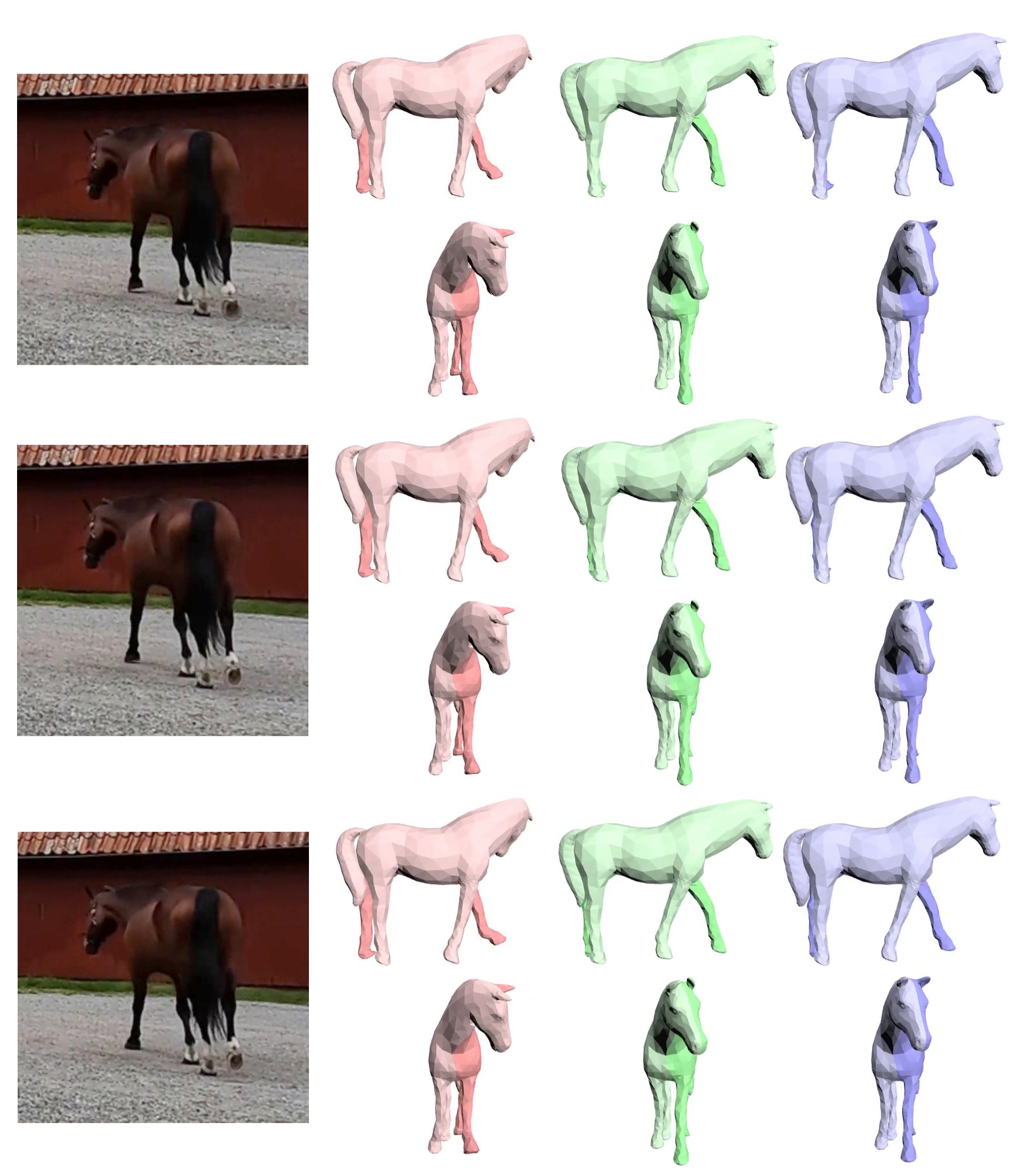}
      \subcaption{Self-occlusion.}
    \label{fig:example_H3S1_1_self_occlusion}
    \end{subfigure}}
    \hspace{0.2mm}
{\begin{subfigure}[b]{0.32\textwidth}
	\centering
     \includegraphics[width=0.97\linewidth]{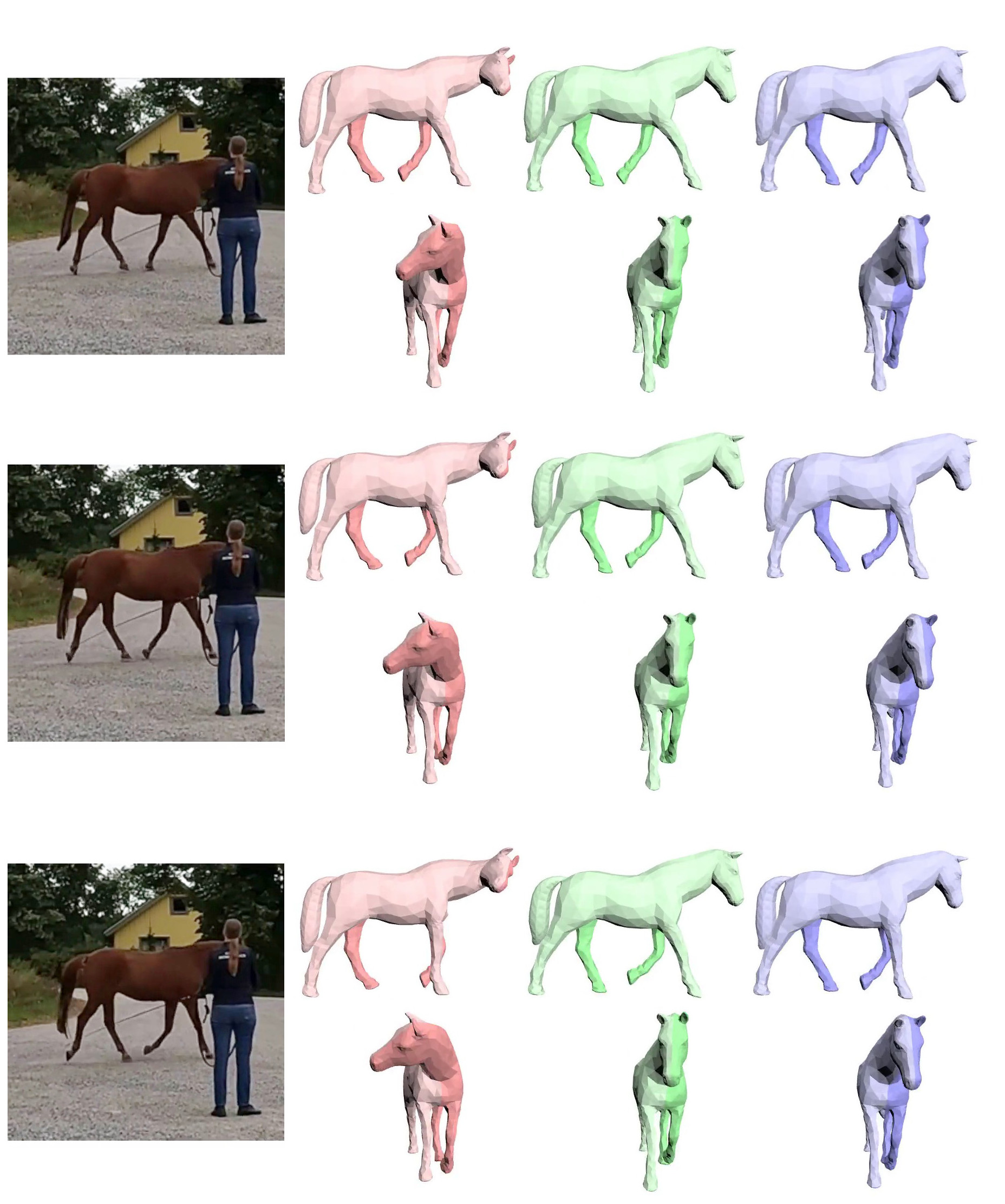}
      \subcaption{The head of the horse is occluded.}
    \label{fig:example_H4S1_4_occluded}
    \end{subfigure}}
    \hspace{0.2mm}
{\begin{subfigure}[b]{0.32\textwidth}
	\centering
     \includegraphics[width=0.97\linewidth]{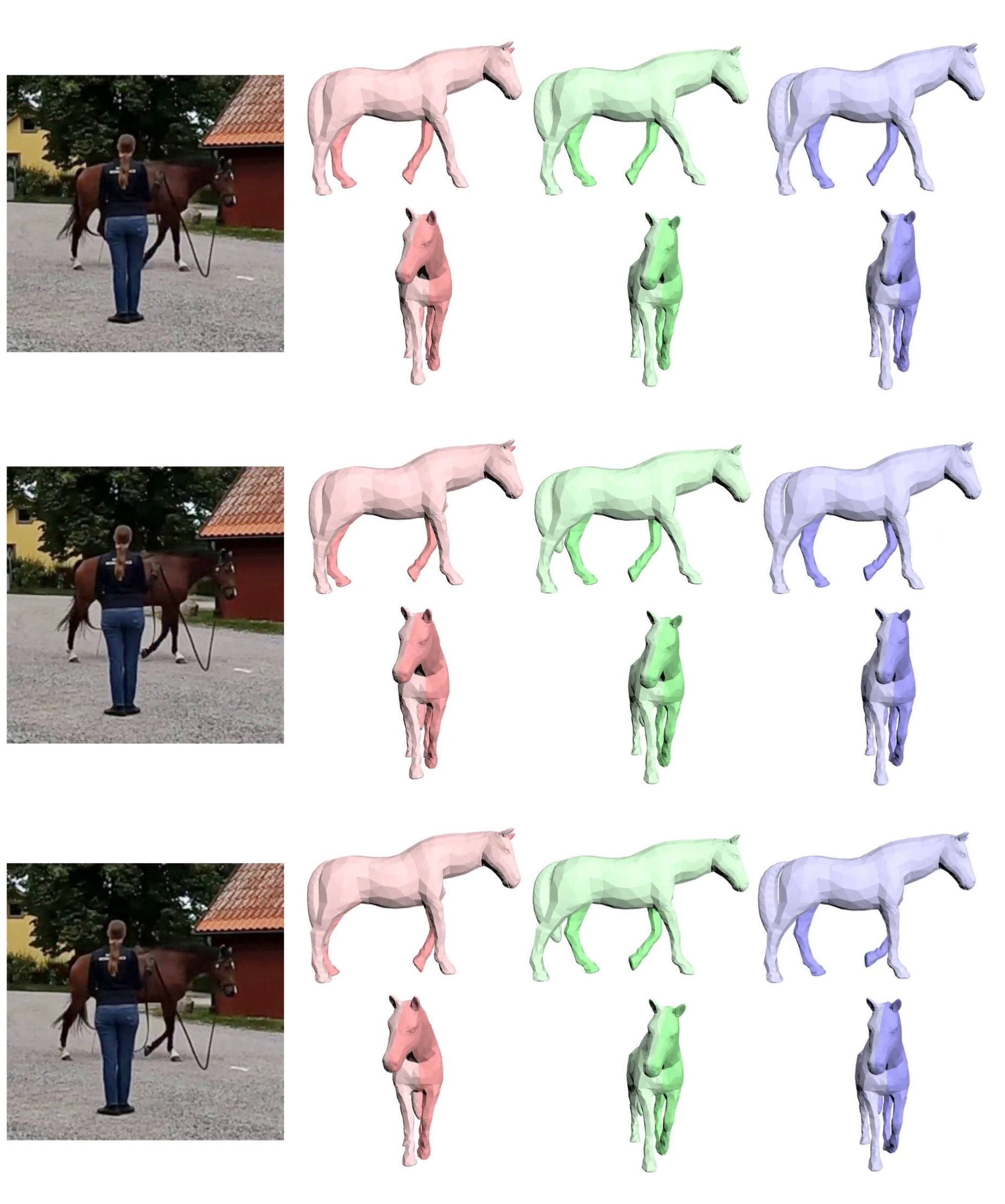} 
      \subcaption{The left hind leg is occluded.}
    \label{fig:example_H3S1_4_occluded}
    \end{subfigure}}
    \vspace{-3mm}
    \caption{Sample outputs on the Outdoor Dataset. Image-only Network (in \textcolor{pink}{LightRed}), Early-fusion Network (in \textcolor{green}{LightGreen}), Model-fusion Network (in \textcolor{blue}{LightBlue}).
    }
    \label{fig:outdoor_results1}
    \vspace{-5mm}
\end{figure*}

\vspace{-5mm}
\paragraph{Experiments on \outdoor{}}

We demonstrate the results on the \outdoor{}. 
The \outdoor{} poses greater challenges due to varied lighting and horses' free movements. We consider two strategies: Per-horse basis training, using 80\% of videos for training and 20\% for testing per horse; Inter-horse training, dividing horses into Test Data 1 and Test Data 2 based on appearance similarity. 
\begin{table}[b]
\centering
\vspace{-5mm}
\caption{Evaluation on the Outdoor dataset. $^*$ denotes networks with data augmentation. ``A / B'' denotes results with original data A and synthetic occluder B.}\label{tab:inter-horse}
\vspace{-2mm}
\setlength{\tabcolsep}{3pt}
\scalebox{0.85}{
\begin{tabular}{c|cc|cc} 
\toprule
\multirow{2}{*}{Network}  & \multicolumn{2}{c|}{Test Data 1 (A / B)} & \multicolumn{2}{c}{Test Data 2 (A / B)} \\ 
\cline{2-5}
 & IOU($\uparrow$) & PCK@0.1($\uparrow$) & IOU($\uparrow$) & PCK@0.1($\uparrow$) \\ 
\midrule
Image-only & \textbf{0.70} / \textbf{0.60} & 0.95 / 0.77 & 0.47 / \textbf{0.43} & 0.56 / 0.43 \\
Early-fusion & 0.69 / 0.59 & \textbf{0.96} / 0.77 & 0.42 / 0.38 & 0.46 / 0.37 \\
Model-fusion & 0.69 / 0.59 & \textbf{0.96} / \textbf{0.78} & \textbf{0.51} / \textbf{0.43} & \textbf{0.64} / \textbf{0.44} \\
\midrule
Image-only$^*$ & \textbf{0.70} / 0.59 & \textbf{0.95} / 0.76 & 0.57 / 0.51 & 0.78 / 0.59 \\
Early-fusion$^*$ & 0.69 / 0.58 & \textbf{0.95} / 0.76 & 0.57 / 0.48 & 0.78 / 0.56 \\
Model-fusion$^*$ & \textbf{0.70} / \textbf{0.60} & \textbf{0.95} / \textbf{0.77} & \textbf{0.61} / \textbf{0.53} & \textbf{0.81} / \textbf{0.64} \\
\bottomrule
\end{tabular}
}
\end{table}

Sample outputs for per-horse training demonstrate that all networks perform similarly when horses are fully visible (Fig.~\ref{fig:example_H3S1_1_full_visible}), but differ under occlusion. Self-occlusion cases (Fig.\ref{fig:example_H3S1_1_self_occlusion}) show the Early- and \ModelFusion{} networks outperforming the \Imageonly{} Network, especially in front leg and head pose estimation, indicating that the training with audio integration enhances the pose estimation. 
For human-induced occlusions (e.g., head occlusion Fig.\ref{fig:example_H4S1_4_occluded} and leg occlusion \ref{fig:example_H3S1_4_occluded}), the \Imageonly{} network struggles with the prediction of head poses (Fig.\ \ref{fig:example_H4S1_4_occluded}) and left hind leg poses (Fig.\ \ref{fig:example_H3S1_4_occluded})), in contrast to the fusion networks which leverage audio cues to improve natural neck and hind leg pose estimations, respectively. The enhanced head and leg poses predicted by the fusion networks can be due to the specific movement pattern of horses. The networks learn the correlation between sound and leg movements, and since leg motion directly influences head position, the integration of audio allows for more natural predictions of both head and leg poses. This results in more accurate pose estimations, even in the presence of occlusions. More examples are in \textit{Supplementary Material}.  

Under the Per-horse basis training approach, a 2D evaluation using PCK and IOU metrics is less informative, due to the predominance of strong image cues. Furthermore, in the case of Inter-horse training, we present a 2D analysis employing IOU on full frame and PCK based on the pseudo-ground truth, covering both the original dataset and data with synthetic occluder, as detailed in Tab.~\ref{tab:inter-horse}. 
The results on the original dataset show similar performance across networks for Test Data 1. 
The \ModelFusion{} network excels in Test Data 2, outperforming the \EarlyFusion{} network. The results of introduced color jittering during training show that data augmentation reduces the domain gap, notably enhancing \ModelFusion{}'s performance on Test Data 2. These findings highlight the \ModelFusion{} network better handles the noisy visual cues from corrupted test data.
\vspace{-2mm}
\section{Conclusions}
\vspace{-1mm}
In this study, we investigate using 
both audio and monocular video for 3D horse reconstruction.
We adopt the hSMAL model to represent the 3D articulated horse, and introduce two strategies for audio-video fusion: Early fusion, where audio and video features are concatenated in the first stages of the network, and Model fusion which leverages audio information only during the training phase. 
Our fusion models achieve more accurate reconstructions and natural poses, even with appearance shifts or visual ambiguities, which indicates the advantage of combining audio and video data for enhanced 3D pose estimation.
\vspace{-4mm}
\paragraph{Current limitations and future work.} We capture audio primarily of ground contact with a fixed camera. Future efforts will include attaching a microphone to the horse to capture breathing and other body sounds that are independent of the type of ground. We also aim to apply this method to other species like dogs, which also produce breathing sounds while moving.

\clearpage
\setcounter{page}{1}
\maketitlesupplementary

\section{Dataset}
In Section 4 of the main paper, we describe two datasets for our experiments. We here provide more detailed information about two datasets.
\vspace{-5mm}
\paragraph{\treadmill{}}
The Horse \treadmill{}, acquired from the University of Z{\"u}rich~\cite{rhodin2018vertical}, 
includes recordings of ten horse subjects trotting on a treadmill. Due to camera calibration problems, we exclude three subjects, focusing on the remaining seven: one white and six dark-colored (brown or black) horses, yielding a total of 702.24 seconds of video at 25 fps. This dataset is unique in offering synchronized video, audio, and 3D motion capture data. 

We map 3D motion capture to images for 34 ground-truth keypoints, add four tail keypoints with DeepLabCut~\cite{Mathisetal2018,NathMathisetal2019}, and generate segmentations via OSVOS~\cite{caelles2017one}. Audio is denoised using Aukit~\cite{aukit}. With horses centered in frames, we assume the bounding box is the full image, and resize each frame to $224 \times 224$.
\vspace{-5mm}
\paragraph{Outdoor Dataset}\label{sec:outdoor}
To complement the controlled conditions of the Treadmill Dataset, we create the \outdoor{} to assess our network's performance in a more natural setting. Captured with a GoPro10 camera at 4K resolution and synchronized audio, this dataset includes four horses of diverse colors (white, black, brown, red) and sizes. They perform walk, trot, and canter motions in both clockwise and anti-clockwise directions, under human guidance via a line attached to their head collars, amounting to 1604.54 seconds of video recordings at 30 fps.

Here we use Detectron2~\cite{wu2019detectron2} to obtain horse silhouettes and their bounding boxes $G_{SIL}$. ViTPose+~\cite{xu_vitpose_2022} provides 17 2D pseudo-ground-truth key points, with a confidence threshold of 0.5, from which we derive a keypoint-based bounding box $G_{KP}$. A few frames where the detector fails are manually labeled. The final bounding box $G$ combines both keypoint and silhouette data for enhanced accuracy.

\vspace{-2mm}
\begin{scriptsize}
\begin{equation}
G = 
\begin{cases}
G_{KP}, & IoU(M_{SIL},M_{KP}) = 0, \\
G_{KP}, & \frac{M_{KP}}{M_{SIL}} > \gamma, \\
G_{SIL} \cup G_{KP}, & \text{otherwise},
\end{cases}
\end{equation}
\end{scriptsize}

where $M_{SIL}$ and $M_{KP}$ are the pixel area of $G_{SIL}$ and $G_{KP}$, respectively. $\gamma$ is a preset threshold, that we set at 2.78. 
Then, the bounding box images are resized to $224 \times 224$ pixels.
The silhouette loss is ignored for frames where the final bounding box is not calculated with $G_{SIL}$.

For each 2D keypoint $K_i$, we define a corresponding 3D point $\kappa_i$.
In the case of the \treadmill{}, where 2D keypoints $K$ are projected from the mocap data, we manually select a point on the horse model surface and express it with barycentric coordinates of the neighboring vertices.
When the 2D keypoints $K$ are obtained from DeepLabCut or ViTPose+, we define 3D keypoints as the interpolation of a set of model vertices, such that the keypoints can also represent skeleton joints.

\section{Network Details}
In Section 3 of the main paper, we describe the network architectures. Here we provide more information about the training loss and the implementation details. 
\vspace{-5mm}
\paragraph{Training loss}
Both regression networks are trained in an end-to-end fashion. 
The training loss is defined as:
\vspace{-2mm}
\begin{equation}
L =  L_{KP} + L_{SIL} + L_{SMOOTH} + L_{HSMAL} ~. 
\end{equation}

The 2D keypoint loss $L_{KP}$ is defined as:
\begin{footnotesize}
\begin{equation}
\begin{aligned}
L_{KP} &= \omega_{KP} \frac{\sum^{T} \sum_{j=1}^{J} \lambda_{j}^2 \rho (\left\| \tilde{K}^{{full}^{j}}_{t} - K^{{full}^{j}}_{t} \right\|_{2}) }{\sum^{T} \sum_{j=1}^{J} \lambda_{j}^2}, \\
\tilde{K}^{{full}^{j}}_{t}&=\prod(\kappa^{j}_{t} , \Omega^{full}_{t}),
\end{aligned}
\end{equation}
\end{footnotesize}
where $\kappa_t$ are the body keypoints on the model, projected on the full image as {\small $\tilde{K}^{full}_{t}$} with perspective projection $\prod$. $K^{full}_t$ are the ground-truth 2D keypoints with confidence scores $\lambda$, and $\rho$ is the Geman-McClure robustifier~\cite{geman1987statistical}.

To constrain the model shape, we use segmentation for supervision. A silhouette loss $L_{SIL}$ is defined as the smooth L1 loss between the projected model silhouette $\tilde{S}^{crop}_{1:T}$ and the ground-truth silhouette $S^{crop}_{1:T}$: 

\begin{footnotesize}
\begin{equation}
\begin{aligned}
L_{SIL} &= \omega_{SIL} \sum^{T} smoothL1( \  \tilde{S}^{crop}_{t} , S^{crop}_{t} ),\\
\tilde{S}^{crop}_{t}&=\xi ( \mathbf{v_t} , \Omega^{crop}_{t} ) ~,
\end{aligned}
\end{equation}
\end{footnotesize}
where $\xi$ is the Pytorch3D function that renders the model silhouette in the cropped images. 

To increase the temporal smoothness of the prediction, we use a smoothness loss $L_{SMOOTH}$ \cite{loper2014mosh,Zhang:ICCV:2021}, to penalize the difference between consecutive frames, defined as:
\begin{footnotesize}
\begin{equation}
L_{SMOOTH} = \omega \frac{1}{N(T-2)}\\
~\sum_{t=3}^{T} \left \| \chi_t - 2\chi_{t-1} + \chi_{t-2} \right \|^{2} ,
\end{equation}
\end{footnotesize}
where $N$ is the length of the input data. $\chi$ represents different predictions, namely the predicted pose parameters and the global rotation in rotation matrix representation, ({\small $R(\theta^{Joints}_{1:T})$} and {\small $R(\theta^{Global}_{1:T})$}), or the full translation of the original cameras {\small $\Gamma^{full}_{1:T}$}, with corresponding weights {\small $\omega^{\theta^{Joints}}_{SMOOTH}$},   {\small $\omega^{\theta^{Global}}_{SMOOTH}$},  {\small $\omega^{\Gamma^{full}}_{SMOOTH}$}.

The prior loss $L_{HSMAL}$ is the weighted sum of the shape and the pose priors of the hSMAL model, defined in \cite{li2021hsmal}, with corresponding weights $\omega^{\beta}_{Prior}$ and $\omega^{\theta}_{Prior}$.

\vspace{-5mm}
\paragraph{Training Detail}
We use a ResNet-50 backbone network to extract visual and audio features. 
The Temporal Encoder, adopted from \cite{humanMotionKanazawa19}, consists of a residual block with two group norm layers with 32 groups and two 1D convolutional layers, with a filter size of 2080. For the input to the temporal encoder, we concatenate the image features from ResNet with bounding box information per frame, where the bounding box has been padded to the length of 32. Following the residual block, the data is processed through a fully-connected layer to get the final visual input features with a dimension of 2048. Our analysis operates on video segments spanning $T=5$ frames.

We train the networks with a learning rate of $5 \times 10^{-5}$ until the training stabilizes, corresponding to 300 epochs for the Treadmill Dataset and 500 epochs for the Outdoor Dataset. We set the weights for each loss as:  
\vspace{-3mm}
\begin{eqnarray}
\omega_{KP}&=&0.001,\nonumber\\
\omega_{SIL}&=&1\times10^{-4},\nonumber\\
\omega^{\beta}_{Prior}&=&50,\nonumber\\  \omega^{\theta}_{Prior}&=&0.01,\nonumber\\ \omega^{\Gamma^{full}}_{SMOOTH}&=&0.1,\nonumber\\ 
\omega^{\theta^{Global}}_{SMOOTH}&=&0.2, \nonumber\\ \omega^{\theta^{Joints}}_{SMOOTH}&=&10.\nonumber
\end{eqnarray}
We assume that both modalities in the \ModelFusion{} Network contribute equally to the pose estimation, setting the equal weight to the pose estimated from audio and video channels. 

For inference, we choose the model with the lowest loss on the validation set. We set a sliding window to select overlapping clips for each test video and consider the result from the middle frame in each clip.  

For network parameters, the \Imageonly, \EarlyFusion, \ModelFusion networks have 98 million, 134 million, and 121 million training parameters, respectively. In the Treadmill dataset, these networks are trained using three 2080TIs with a batch size of 9, and take 13 hours,17 hours, 24 hours, respectively. For testing on a single 2080TI with a batch size of 1, they take 27 minutes, 27 minutes, 54 minutes, respectively.

\section{Synthetic Occlusions}

In Section 4 of the main paper, we describe one of the experiments with adding artificial visual noise to the original data. The purpose of this is to demonstrate the robustness of our models in the presence of visual noise by adding a synthetic occluder to the images. 

\begin{figure}[h]
\vspace{-2mm}
  \centering
  \begin{subfigure}{0.2\textwidth} 
    \centering
    \includegraphics[width=1.\linewidth]{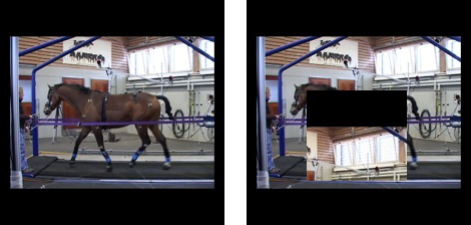}
    \caption{Test Data 1.}
  \end{subfigure}
  \hfill
  \begin{subfigure}{0.2\textwidth} 
    \centering
    \includegraphics[width=1.\linewidth]{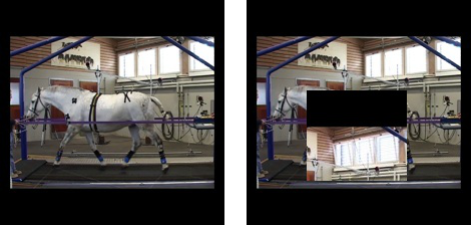}
    \caption{Test Data 2.}
  \end{subfigure}
  \vspace{-3mm}
  \caption{Simulation of visual ambiguity at test time in the Treadmill Dataset. \textbf{Left:} the original frame. \textbf{Right:} the frame corrupted with an occluder.}
  \label{fig:corrupted_treadmill}
  \vspace{-3mm}
\end{figure}

\begin{figure}[h]
\vspace{-2mm}
  \centering
  \begin{subfigure}{0.2\textwidth}
    \centering
    \includegraphics[width=1.0\linewidth]{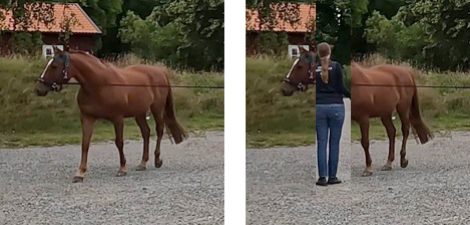}
    \caption{Test Data 1.}
  \end{subfigure}
  \hfill
  \begin{subfigure}{0.2\textwidth}
    \centering
    \includegraphics[width=1.0\linewidth]{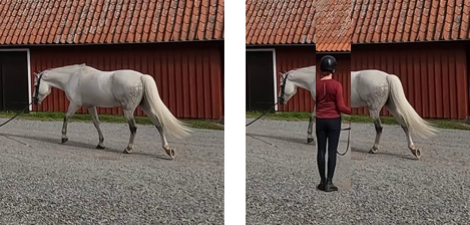}
    \caption{Test Data 2.}
  \end{subfigure}
  \vspace{-3mm}
  \caption{Simulation of visual ambiguity at test time in the Outdoor Dataset. \textbf{Left:} the original frame. \textbf{Right:} the frame corrupted with an occluder.} 
  \label{fig:corrupted}
  \vspace{-3mm}
\end{figure}

In the \treadmill{}, the synthetic occluder is part of the image from the training dataset and covers most areas of the horse. In the \outdoor{}, the synthetic occluder is a human. Some examples of the synthetic occlusions are shown in Fig.\ \ref{fig:corrupted_treadmill} and Fig.\ \ref{fig:corrupted}.


\section{More Qualitative Results on the Outdoor Dataset}
In addition to the quantitative experiments in Per-horse basis training in Section 4 of the main paper, we here provide more qualitative results.

\begin{figure*}[ht]
    \centering
\includegraphics[width=0.98\linewidth]{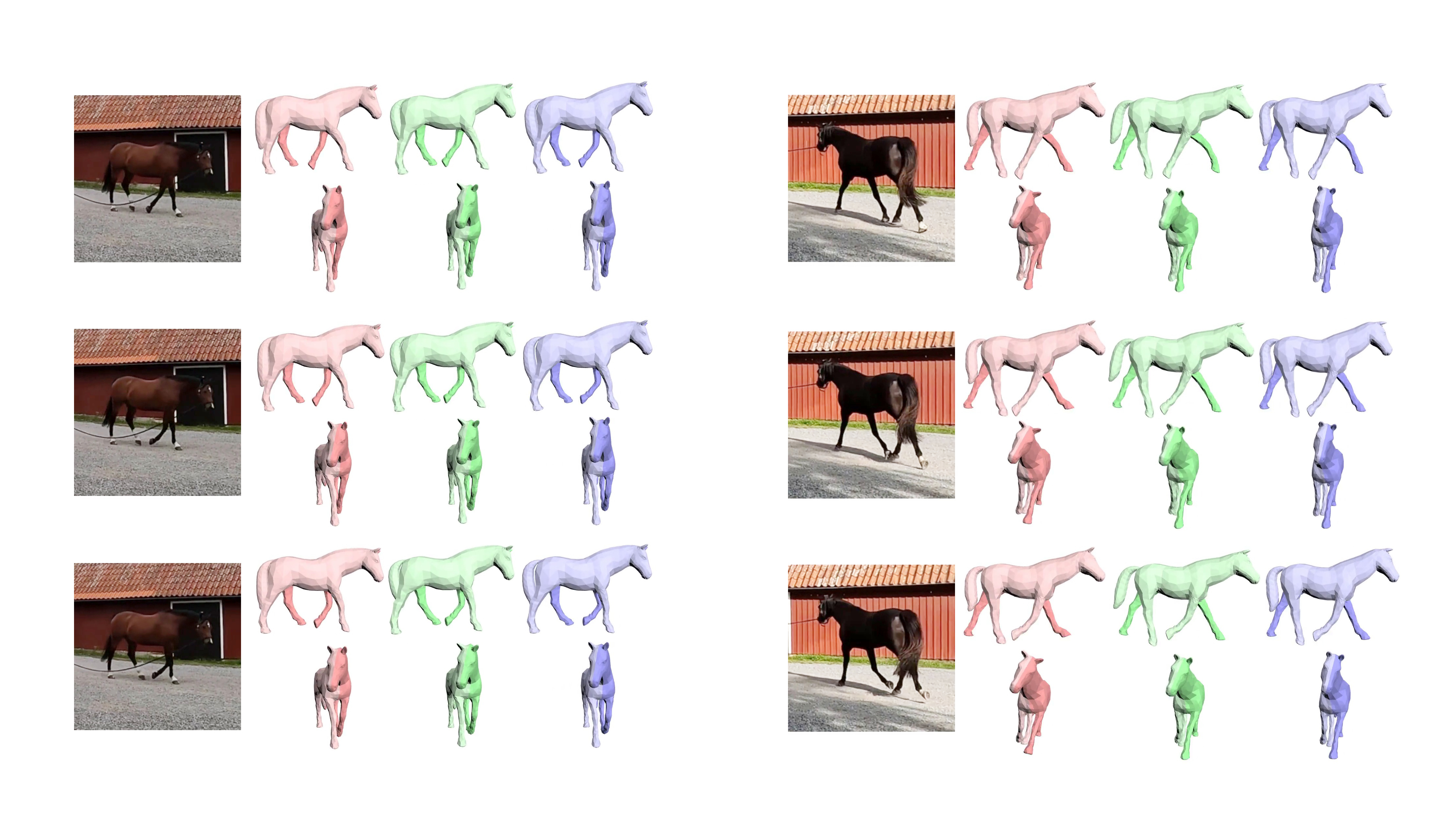}
    \caption{Sample outputs of full body visible on the Outdoor Dataset. Image-only Network (in \textcolor{pink}{LightRed}), \EarlyFusion{} Network (in \textcolor{green}{LightGreen}), \ModelFusion{} Network (in \textcolor{blue}{LightBlue}).
    }
     \vspace{-4mm}
\label{fig:outdoor_full_body_visible_results}
\end{figure*}

\begin{figure*}[ht]
    \centering
\includegraphics[width=0.98\linewidth]{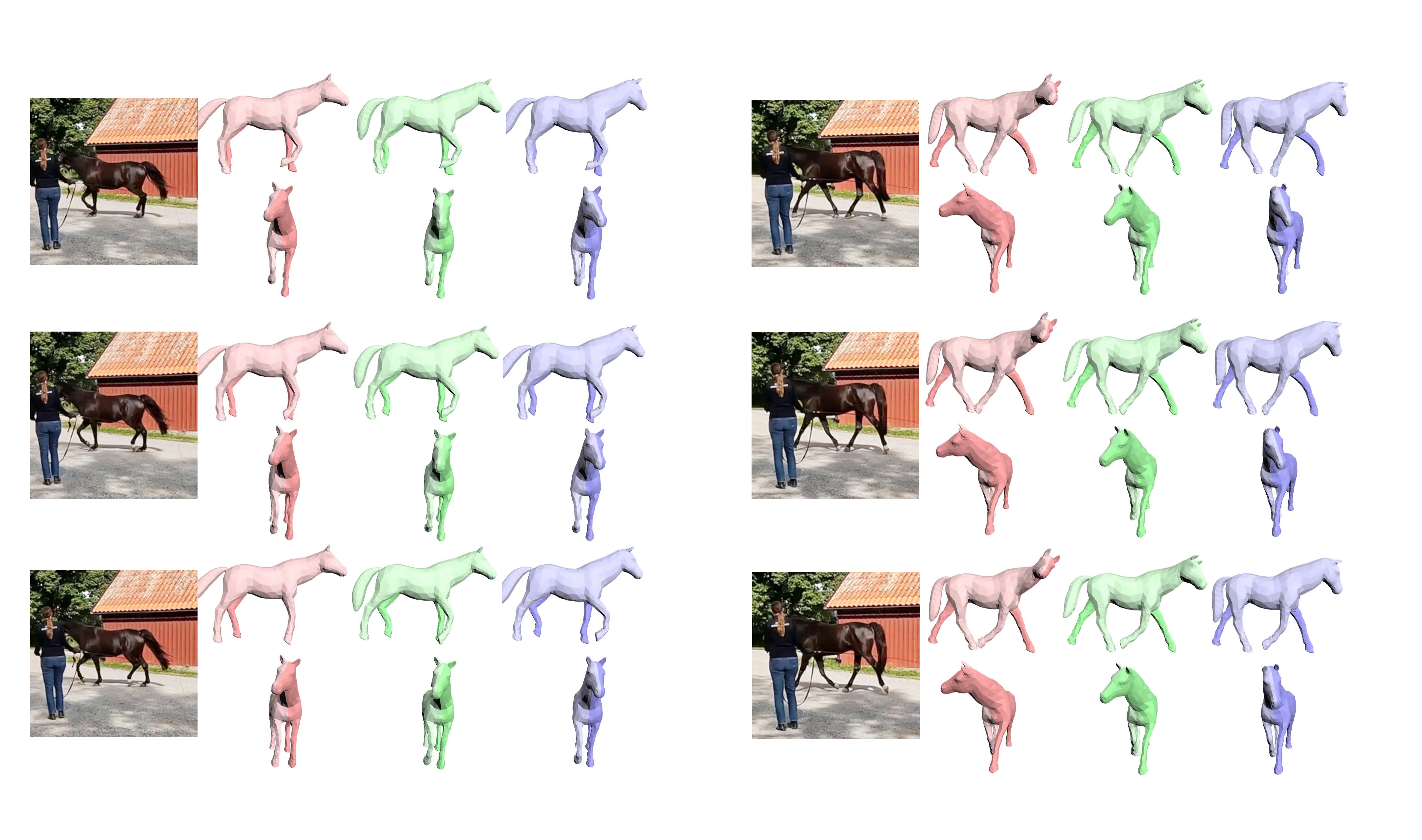}
    \caption{Sample outputs of occlusion on the Outdoor Dataset. Image-only Network (in \textcolor{pink}{LightRed}), \EarlyFusion{} Network (in \textcolor{green}{LightGreen}), \ModelFusion{} Network (in \textcolor{blue}{LightBlue}).
    } \vspace{-4mm}
    \label{fig:outdoor_results3}
\end{figure*}

When the horse is fully visible, the three networks produce very similar pose estimates (in Fig.\ \ref{fig:outdoor_full_body_visible_results}).
Two examples are shown in the case where the horse is occluded by a human in Fig.\ \ref{fig:outdoor_results3}. The Image-only network produces rigid front legs (left) and unnatural head poses (right) that point to the right side, while the Early- and Model-fusion networks predict more natural front legs and neck poses. 
This demonstrates that networks trained with both audio and visual information have the potential to be more robust to occlusion.

 \vspace{-4mm}

\clearpage
{
    \small
    \bibliographystyle{ieeenat_fullname}
    \bibliography{egbib} 
}
\end{document}